\documentclass[twoside]{article}
\PassOptionsToPackage{numbers, compress}{natbib}
\usepackage[accepted]{aistats2023}

\usepackage{algorithm}
\usepackage{algorithmic}

\usepackage{nicefrac}
\usepackage{microtype}
\usepackage{graphicx}
\usepackage{caption}
\usepackage{subcaption}
\usepackage{tikz-cd}
\usetikzlibrary{arrows}
\usepackage{comment}
\usepackage{amssymb}
\usepackage{tikz}
\usepackage{tikz-qtree,tikz-qtree-compat}
\usetikzlibrary{positioning}

\usepackage{amsmath}
\usepackage{amsthm}
\usepackage{tikz}
\usepackage{tikz-qtree,tikz-qtree-compat}

\makeatletter
\newtheorem*{rep@theorem}{\rep@title}
\newcommand{\newreptheorem}[2]{%
\newenvironment{rep#1}[1]{%
 \def\rep@title{#2 \ref{##1}}%
 \begin{rep@theorem}}%
 {\end{rep@theorem}}}
\makeatother

\newtheorem{theorem}{Theorem}
\newreptheorem{theorem}{Theorem}
\newtheorem{corollary}[theorem]{Corollary}
\newreptheorem{corollary}{Corollary}

\usetikzlibrary{positioning}
\newtheorem{definition}[theorem]{Definition}

\usepackage{stackengine}
\def\delequal{\mathrel{\ensurestackMath{\stackon[1pt]{=}{\scriptstyle\Delta}}}}
\begin{document}

%

%

\twocolumn[

\aistatstitle{Probabilities of Causation: Role of Observational Data}

\aistatsauthor{ Ang Li \And Judea Pearl }

\aistatsaddress{   Cognitive Systems Laboratory, \\Department of Computer Science,\\
  University of California Los Angeles\\
  Los Angeles, CA 90095 \\
  \texttt{angli@cs.ucla.edu} \\ \And Cognitive Systems Laboratory, \\Department of Computer Science,\\
  University of California Los Angeles\\
  Los Angeles, CA 90095 \\
  \texttt{judea@cs.ucla.edu} \\} ]

\begin{abstract}
  Probabilities of causation play a crucial role in modern decision-making. Pearl defined three binary probabilities of causation, the probability of necessity and sufficiency (PNS), the probability of sufficiency (PS), and the probability of necessity (PN). These probabilities were then bounded by Tian and Pearl using a combination of experimental and observational data. However, observational data are not always available in practice; in such a case, Tian and Pearl's Theorem provided valid but less effective bounds using pure experimental data. In this paper, we discuss the conditions that observational data are worth considering to improve the quality of the bounds. More specifically, we defined the expected improvement of the bounds by assuming the observational distributions are uniformly distributed on their feasible interval. We further applied the proposed theorems to the unit selection problem defined by Li and Pearl.
\end{abstract}

\section{INTRODUCTION}
Probabilities of causation are widely used in social science, political science, health science, economics, etc. For example, Li and Pearl used a linear combination of probabilities of causation to solve the unit selection problem \cite{li:pea22-r521, li:pea22-r517, li:pea19-r488, li2022unit}, showing the advantage of their model over the A/B test heuristic. Also, Scott and Pearl showed that the probabilities of causation should be considered in personal decision-making \cite{mueller:pea-r513}. In addition, probabilities of causation have improved the accuracy of machine learning models by including probabilities of causation terms in the label of training data \cite{li2020training}. Machine learning models can also be used to learn the bounds of Probabilities of causation if the proper training data are provided \cite{li:pea22-r519, li:pea22-r520}.

The probabilities of causation have been studied for decades. Pearl first defined three binary probabilities of causation (i.e., PNS, PN, and PS) \cite{pearl1999probabilities} using the structural causal model (SCM) \cite{galles1998axiomatic,halpern2000axiomatizing}. The sharp bounds of these probabilities of causation are then derived by Tian and Pearl (referred to as Tian-Pearl's Theorem) \cite{tian2000probabilities} utilizing a combination of experimental and observational data, and Balke's linear programming \cite{balke1997probabilistic}. These bounds are then improved with additional covariate information and causal structures \cite{dawid2017, li2022bounds, pearl:etal21-r505}. Besides, the theoretical foundation of non-binary probabilities of causation was studied by Li and Pearl \cite{li:pea-r516}.

The probabilities of causation are not generally identifiable (i.e., no point estimation exists if there is no additional assumption); therefore, all the works mentioned above focused on bounding the probabilities of causation using a combination of experimental and observational data. However, observational data are not always available in the real world, and observational studies usually take longer time to conduct. For example, a new drug is generally proven to the market by only a randomized controlled clinical study (i.e., experimental study), lacking observational study (i.e., patients are free to choose whether or not to take the drug). In such a situation, Tian-Pearl's Theorem can still provide valid bounds of the probabilities of causation using pure experimental data. However, the bounds might be less informative (e.g., $0.1\le PNS\le 0.9$). We then wonder whether we should collect observational data to improve the bounds and what the expected improvement could be. In this paper, we formally defined the expected improvement of lower and upper bounds assuming the observational distributions are uniformly distributed on their feasible interval.

\section{PRELIMINARIES}
Here, we review the three binary probabilities of causation defined by Pearl \cite{pearl1999probabilities} and their sharp bounds derived by Tian and Pearl \cite{tian2000probabilities}. Readers who are familiar with the model may skip this section.

The language of SCM \cite{galles1998axiomatic,halpern2000axiomatizing} will be used in which the probabilities of causation are well-defined. The basic counterfactual sentence ``Variable $Y$ would have the value $y$, had $X$ been $x$'' is denoted by $Y_{X=x}=y$ and shorted as $y_x$. The experimental data are in the form of the causal effects $P(y_x)$, and the observational data are in the form of a joint probability function $P(x, y)$.

The three binary probabilities are then defined as follows:

\begin{definition}[Probability of necessity (PN)]
Let $X$ and $Y$ be two binary variables in a causal model $M$, let $x$ and $y$ stand for the propositions $X=true$ and $Y=true$, respectively, and $x'$ and $y'$ for their complements. The probability of necessity is defined as the expression \cite{pearl1999probabilities}\\
\begin{eqnarray}
\text{PN} &\delequal& P(Y_{x'}=false|X=true,Y=true)\nonumber\\
 &\delequal&  P(y'_{x'}|x,y) \nonumber
\label{pn}
\end{eqnarray}
\end{definition}

\begin{definition}[Probability of sufficiency (PS)]
Let $X$ and $Y$ be two binary variables in a causal model $M$, let $x$ and $y$ stand for the propositions $X=true$ and $Y=true$, respectively, and $x'$ and $y'$ for their complements. The probability of sufficiency is defined as the expression \cite{pearl1999probabilities}\\
\begin{eqnarray}
\text{PS} \delequal P(y_x|y',x') \nonumber
\label{ps}
\end{eqnarray}
\end{definition}

\begin{definition}[Probability of necessity and sufficiency (PNS)] Let $X$ and $Y$ be two binary variables in a causal model $M$, let $x$ and $y$ stand for the propositions $X=true$ and $Y=true$, respectively, and $x'$ and $y'$ for their complements. The probability of necessity and sufficiency is defined as the expression \cite{pearl1999probabilities}\\
\begin{eqnarray}
\text{PNS}\delequal P(y_x,y'_{x'}) \nonumber
\label{pns}
\end{eqnarray}
\end{definition}

The bounds of the above probabilities of causation are as follows:
\begin{eqnarray}
\text{PNS} \ge \max \left \{
\begin{array}{cc}
0, \\
P(y_x) - P(y_{x'}), \\
P(y) - P(y_{x'}), \\
P(y_x) - P(y)
\end{array}
\right \}
\label{pnslb}
\end{eqnarray}

\begin{eqnarray}
\text{PNS} \le \min \left \{
\begin{array}{cc}
 P(y_x), \\
 P(y'_{x'}), \\
P(x,y) + P(x',y'), \\
P(y_x) - P(y_{x'}) +\\
+ P(x, y') + P(x', y)
\end{array} 
\right \}
\label{pnsub}
\end{eqnarray}

\begin{eqnarray*}
\text{PN} \ge \max \left \{
\begin{array}{cc}
0, \\
\frac{P(y)-P(y_{x'})}{P(x,y)}
\end{array} 
\right \}
\label{pnlb}
\end{eqnarray*}

\begin{eqnarray*}
\text{PN} \le
\min \left \{
\begin{array}{cc}
1, \\
\frac{P(y'_{x'})-P(x',y')}{P(x,y)}
\end{array}
\right \}
\label{pnubb}
\end{eqnarray*}

\begin{eqnarray*}
\text{PS} \ge \max \left \{
\begin{array}{cc}
0, \\
\frac{P(y')-P(y'_{x})}{P(x',y')}
\end{array} 
\right \}
\label{pnlbb}
\end{eqnarray*}

\begin{eqnarray*}
\text{PS} \le
\min \left \{
\begin{array}{cc}
1, \\
\frac{P(y_{x})-P(x,y)}{P(x',y')}
\end{array}
\right \}
\label{pnub}
\end{eqnarray*}

The theoretical proof of these bounds can be found in \cite{li:pea19-r488}. Note that in these bounds, both experimental data (e.g., $P(y_x)$ in Equation \ref{pnslb}) and observational data (e.g., $P(y)$ in Equation \ref{pnslb}) are used. If the observational data are not available, the bounds of PNS become the following:

\begin{eqnarray}
\text{PNS} \ge \max \left \{
\begin{array}{cc}
0, \\
P(y_x) - P(y_{x'})
\end{array}
\right \}
\label{pnslbexp}
\end{eqnarray}

\begin{eqnarray}
\text{PNS} \le \min \left \{
\begin{array}{cc}
 P(y_x), \\
 P(y'_{x'})
\end{array} 
\right \}
\label{pnsubexp}
\end{eqnarray}

In this paper, we will focus on the expected improvement of the lower (i.e., Equation \ref{pnslb} $-$ Equation \ref{pnslbexp}) and upper (i.e., Equation \ref{pnsubexp} $-$ Equation \ref{pnsub})  bounds of PNS. The results can be extended to other probabilities of causation.

\section{MAIN RESULTS}
Suppose a decision maker has the experimental data $P(y_x)$ and $P(y_{x'})$, and obtained the bounds of PNS by Equations \ref{pnslbexp} and \ref{pnsubexp}. To improve the obtained bounds, the decision maker wondered whether he should conduct an observational study to obtain observational data (i.e., apply Equations \ref{pnslb} and \ref{pnsub}). We then have the following theorems (the proof of all theorems in this paper is provided in the appendix):
\begin{theorem}
Given experimental data $P(y_x)$ and $P(y_{x'})$ and let $D=P(y)$ be a random variable. Let $L$ be the lower bound of PNS using pure experimental data and $L'$ be the lower bound of PNS using a combination of experimental and observational data. If $D$ is uniformly distributed on its feasible interval $[max(0, P(y_x)-P({y'}_{x'})), min(1, P(y_x)+P(y_{x'}))]$ ,and $P(y_x)+P(y_{x'})\ne 0$ and $P({y'}_x)+P({y'}_{x'})\ne 0$, then we have the expectation of the increased lower bound $E(L'-L)$ as follows:
\begin{eqnarray*}
&&E(L'-L)\\
&=&\frac{\min\{P^2(y_x), P^2({y'}_x), P^2({y}_{x'}), P^2({y'}_{x'})\}}{\min\{P(y_x)+P(y_{x'}),P({y'}_x)+P({y'}_{x'})\}}
\end{eqnarray*}
where,\\
\begin{eqnarray*}
P({y'}_x) = 1 - P(y_x),\\
P({y'}_{x'}) = 1 - P(y_{x'}),
\end{eqnarray*}
\begin{eqnarray*}
\text{L'} = \max \left \{
\begin{array}{cc}
0, \\
P(y_x) - P(y_{x'}), \\
D - P(y_{x'}), \\
P(y_x) - D
\end{array}
\right \},
\end{eqnarray*}
\begin{eqnarray*}
\text{L} = \max \left \{
\begin{array}{cc}
0, \\
P(y_x) - P(y_{x'})
\end{array}
\right \}.
\end{eqnarray*}
\label{thm1}
\end{theorem}

\begin{theorem}
Given experimental data $P(y_x)$ and $P(y_{x'})$ and let $D=P(x,y)+P(x',y')$ be a random variable. Let $U$ be the upper bound of PNS using pure experimental data and $U'$ be the upper bound of PNS using a combination of experimental and observational data. If $D$ is uniformly distributed on its feasible interval $[max(0, P(y_x)-P(y_{x'})), min(1, P(y_x)+ P({y'}_{x'}))]$, and $P(y_x)+P({y'}_{x'})\ne 0$ and $P({y'}_x)+P({y}_{x'})\ne 0$, then we have the expectation of the decreased upper bound $E(U-U')$ as follows:
\begin{eqnarray*}
&&E(U-U')\\
&=&\frac{\min\{P^2(y_x), P^2({y'}_x), P^2({y}_{x'}), P^2({y'}_{x'})\}}{\min\{P(y_x)+P({y'}_{x'}),P({y'}_x)+P({y}_{x'})\}}
\end{eqnarray*}
where,
\begin{eqnarray*}
P({y'}_x) = 1 - P(y_x),\\
P({y'}_{x'}) = 1 - P(y_{x'}),
\end{eqnarray*}
\begin{eqnarray*}
\text{U'} = \min \left \{
\begin{array}{cc}
 P(y_x), \\
 P(y'_{x'}), \\
D, \\
P(y_x) + P({y'}_{x'}) -D
\end{array}
\right \},
\end{eqnarray*}
\begin{eqnarray*}
\text{U} = \min \left \{
\begin{array}{cc}
 P(y_x), \\
 P(y'_{x'})
\end{array}
\right \}.
\end{eqnarray*}
\label{thm2}
\end{theorem}

The expected increased lower bound and decreased upper bound represent the improvement of the bounds of PNS when considering observational data. Note that Theorem \ref{thm1} requires $P(y_x) + P(y_{x'})\ne 0$ and $P({y'}_x) + P({y'}_{x'})\ne 0$, and Theorem \ref{thm2} requires $P(y_x) + P({y'}_{x'})\ne 0$ and $P({y'}_x) + P({y}_{x'})\ne 0$. In fact, if $P(y_x) + P(y_{x'})= 0$ or $P(y_x) + P({y'}_{x'})= 0$ or $P({y'}_x) + P({y'}_{x'})= 0$, PNS is reduced to the point estimation $0$ (i.e., $PNS=0$), and if $P({y'}_x) + P({y}_{x'})= 0$, PNS is reduced to the point estimation $1$ (i.e., $PNS=1$). Besides, the feasible intervals of $D$ in both theorems come from the general relationship between experimental and observational data (i.e., $P(x,y)\le P(y_x)\le 1 - P(x,y')$) proposed by Tian and Pearl \cite{tian2000probabilities}.

\subsection{Visualization of Theorems}
\begin{figure}
\centering
\includegraphics[width=0.49\textwidth]{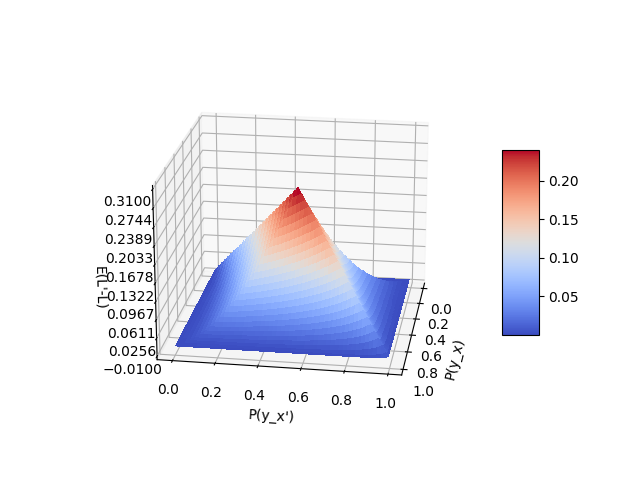}
\caption{$(P(y_x),P(y_{x'}))$ v.s. Expected increased lower bound $E(L'-L)$.}
\label{fig1}
\end{figure}

\begin{figure}
\centering
\includegraphics[width=0.49\textwidth]{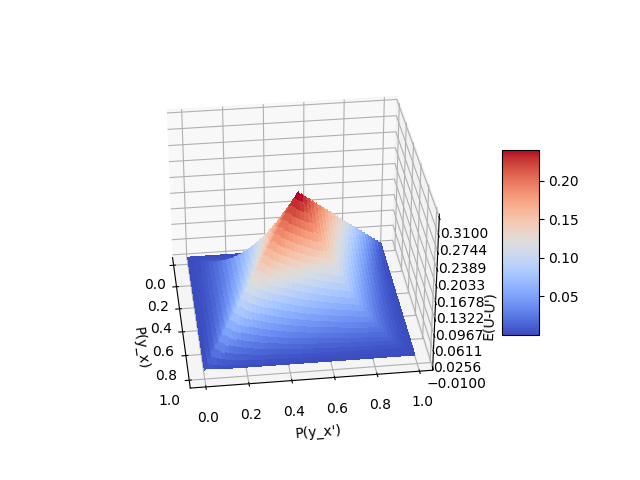}
\caption{$(P(y_x),P(y_{x'}))$ v.s. Expected decreased upper bound $E(U-U')$.}
\label{fig2}
\end{figure}

In order to have a better understanding of Theorems \ref{thm1} and \ref{thm2}. We plotted $(P(y_x),P(y_{x'}))$ v.s. $E$ graphs as shown in Figures \ref{fig1} and \ref{fig2}. Both expected increased lower bounds and decreased upper bounds are high when both $P(y_x)$ and $P(y_{x'})$ are close to $0.5$.

\begin{figure}
\centering
\includegraphics[width=0.49\textwidth]{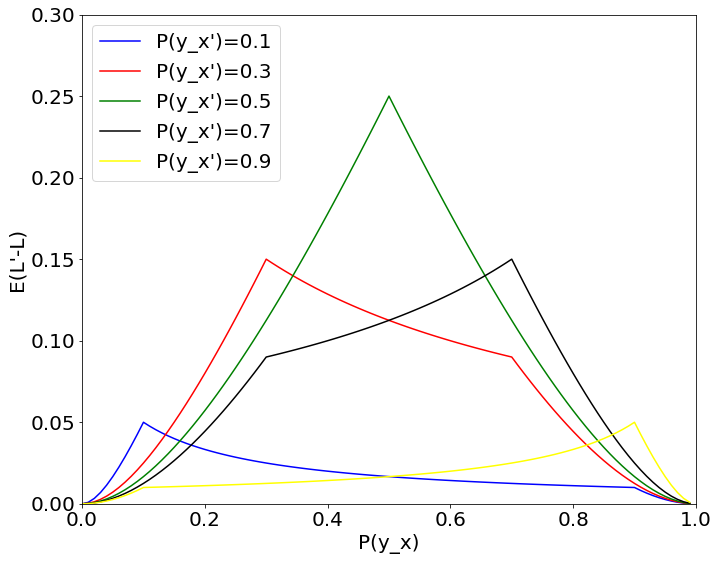}
\caption{$P(y_x)$ v.s. Expected increased lower bound $E(L'-L)$.}
\label{fig3}
\end{figure}

\begin{figure}
\centering
\includegraphics[width=0.49\textwidth]{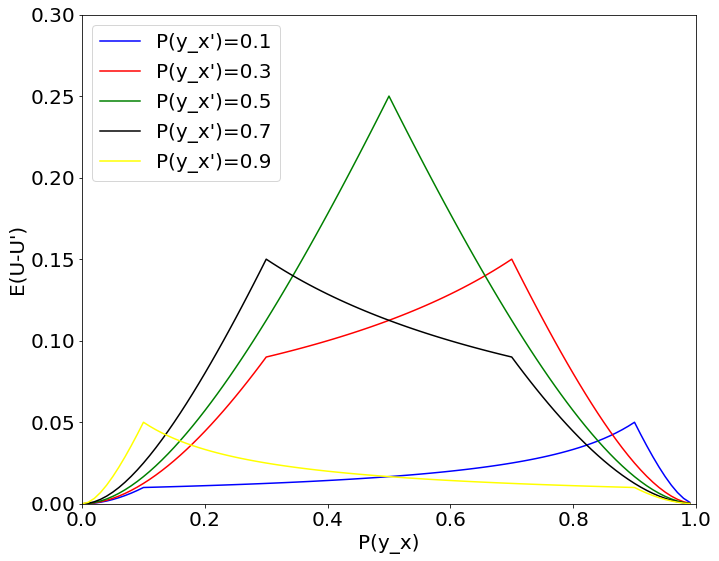}
\caption{$P(y_x)$ v.s. Expected decreased upper bound $E(U-U')$.}
\label{fig4}
\end{figure}

To have a better visualization, we then fixed the value of $P(y_{x'})$ and plotted the $P(y_x)$ v.s. $E$ graphs as shown in Figures \ref{fig3} and \ref{fig4}. For each of the fixed $P(y_{x'})$, the maximum values of expected increased lower bound come from the situation that $P(y_x)=P(y_{x'})$, and the maximum values of expected decreased upper bound come from the situation that $P(y_x)+P(y_{x'})=1$. In addition, same as we concluded from Figures \ref{fig1} and \ref{fig2}, the overall maximum values of both expected increased lower bound and decreased upper bound reached when $P(y_x)=P(y_{x'})=0.5$.


\section{EXAMPLES}
Here, in this section, we provided two simulated examples of how to apply the proposed theorems.
\subsection{PNS of Vaccine}
Consider a pharmaceutical company that invented a new vaccine for a virus. The company wants to claim the effectiveness of the vaccine by showing that the PNS of the vaccine (i.e., the percentage of individuals who would not be affected by the virus if vaccinated and would be affected if unvaccinated) is high.

Let $X=x$ denote that an individual received the vaccine, and $X=x'$ denote that an individual received no vaccine. Let $Y=y$ denote that an individual is not affected by the virus, and $Y=y'$ denote that an individual is affected by the virus.

The pharmaceutical company has conducted an experimental study that $1500$ individuals were forced to receive the vaccine and $1500$ individuals were forced to not (Note that $1500$ is the suggested sample size by Li and Pearl \cite{li:pea22-r518} to have accuracy estimation of PNS). The results are shown in Table \ref{tb1}.

\begin{table}
\centering
\caption{Results of an experimental study where $1500$ individuals were forced to receive the vaccine and $1500$ individuals were forced not to receive the vaccine.}
\begin{tabular}{|c|c|c|}
\hline
&Vaccinated&Unvaccinated\\
\hline
Unaffected&$795$&$720$\\
\hline
Affected&$705$&$780$\\
\hline
\end{tabular}
\label{tb1}
\end{table}

If frequentist is used for experimental data, we have $P(y_x)=0.53$ and $P(y_{x'})=0.48$. We then plugged the experimental data into Equations \ref{pnslbexp} and \ref{pnsubexp}, and obtained that $0.05\le PNS \le 0.52$. It is hard for the pharmaceutical company to claim that the vaccine is effective because PNS can be as low as $0.05$ though the upper bound is $0.52$. Thus, they were considering obtaining observational data.

Before that, they plugged the experimental data in to Theorems \ref{thm1} and \ref{thm2}, and obtained that $E(L'-L)$ (i.e., Expected increased lower bound) is $0.2231$ and $E(U-U')$ (i.e., Expected decreased upper bound) is $0.2325$. The theorem indicates that both lower and upper bounds have non-minor expected improvement. The pharmaceutical company then conducted an observational study with $1500$ individuals who have access to the vaccine, where $660$ individuals chose to receive the vaccine and $840$ individuals chose not to. The results of the observational study are shown in Table \ref{tb2}.

Again, if frequentist is used for observational data, we have $P(x,y)=0.14$, $P(x',y)=0.06$, $P(x,y')=0.3$, and $P(x',y')=0.5$. We then plugged both experimental and observational data into Equations \ref{pnslb} and \ref{pnsub}, and obtained that $0.33\le PNS \le 0.41$. These bounds are tight enough to conclude that the vaccine is effective because at least $33\%$ of individuals are the ones who would be unaffected by the virus if vaccinated and would be affected if unvaccinated.

\begin{table}
\centering
\caption{Results of an observational study with $1500$ individuals who have access to the vaccine, where $660$ individuals chose to receive the vaccine and $840$ individuals chose not to.}
\begin{tabular}{|c|c|c|}
\hline
&Vaccinated&Unvaccinated\\
\hline
Unaffected&$210$&$90$\\
\hline
Affected&$450$&$750$\\
\hline
\end{tabular}
\label{tb2}
\end{table}

\subsection{PNS of Enticement}
A car dealer wants to send an enticement (i.e., discount) to all customers to encourage them to buy a hybrid car. The manager of the car dealer wants to know how many customers are the ones who would buy the hybrid car if they received the discount and would not otherwise. The manager then conducted an experimental study by offering the discount to $1500$ of its customers and offering no discount to another $1500$ of its customers. The experimental results are shown in Table \ref{tb3}.

Let $X=x$ denote that a customer received the discount and $X=x'$ denote that a customer received no discount. Let $Y=y$ denote that a customer purchased the hybrid car and $Y=y'$ denote that a customer did not purchase the hybrid car.

\begin{table}
\centering
\caption{Results of an experimental study where $1500$ customers were forced to receive the discount and $1500$ customers were forced to not.}
\begin{tabular}{|c|c|c|}
\hline
&Received discount&No discount\\
\hline
Purchased&$150$&$1350$\\
\hline
Not purchased&$1350$&$150$\\
\hline
\end{tabular}
\label{tb3}
\end{table}

The experimental data from Table \ref{tb3} are $P(y_x) = 0.1$ and $P(y_{x'})=0.9$. We can then obtained the bounds of PNS from Equations \ref{pnslbexp} and \ref{pnsubexp}, where $0\le PNS \le 0.1$. In order to convince that there do exist customers who would buy the hybrid car if they received the discount and would not otherwise, the lower bound of PNS should be improved by investigating observational data. 

The car manager then applied Theorem \ref{thm1} and obtained that the expectation of the increased lower bound is $E(L'-L)=0.01$. This means that it is hard to improve the lower bound even combining with observational data; therefore, we suggested that it is no need to conduct an observational study, and the manager should consider other approaches (e.g., obtaining the covariate information \cite{pearl:etal21-r505}) to show there exist the desired customers.

\section{APPLICATION TO UNIT SELECTION PROBLEM}
The unit selection problem defined by Li and Pearl \cite{li:pea19-r488} is to identify individuals who have the desired behavior, for example, the individuals who would have a positive effect if treated and would have a negative effect otherwise.

Let $X=x$ denote that an individual received the treatment and $X=x'$ denote that an individual received no treatment. Let $Y=y$ denote that an individual has a positive effect and $Y=y'$ denote that an individual has a negative effect.

According to Li and Pearl, individuals are divided into four response types: complier (i.e., $(y_x,{y'}_{x'})$), always-taker (i.e., $(y_x,{y}_{x'})$), never-taker (i.e., $({y'}_x,{y'}_{x'})$), and defier (i.e., $({y'}_x,{y}_{x'})$). Suppose the payoff of selecting a complier, always-taker, never-taker, and defier is $(\beta, \gamma, \theta, \delta)$, respectively (i.e., the benefit vector). The objective function defined by Li and Pearl (i.e., the benefit function) is then \cite{li:pea19-r488}
\begin{eqnarray*}
f(c) &=&\beta P(y_{x},y'_{x'}|c)+\gamma P(y_{x},y_{x'}|c) +\\
&&+\theta P(y'_{x},y'_{x'}|c)+\delta P(y'_{x},y_{x'}|c).
\end{eqnarray*}
Note that $c$ is the population-specific characteristics and the benefit function is a linear combination of the probabilities of causation; therefore, Li and Pearl derived the tight bounds of the benefit function as follows:
\begin{theorem}
\label{thm3}
Given a causal diagram $G$ and distribution compatible with $G$, let $C$ be a set of variables that does not contain any descendant of $X$ in $G$, then the benefit function $f(c)=\beta P(y_x,y'_{x'}|c)+\gamma P(y_x,y_{x'}|c)+ \theta P(y'_x,y'_{x'}|c) + \delta P(y_{x'},y'_{x}|c)$ is bounded as follows \cite{li:pea19-r488}:
\begin{eqnarray*}
W+\sigma U\le f(c) \le W+\sigma L\text{~~~~~~~~if }\sigma < 0,\\
W+\sigma L\le f(c) \le W+\sigma U\text{~~~~~~~~if }\sigma > 0,
\end{eqnarray*}
where $\sigma, W,L,U$ are given by,
\begin{eqnarray*}
&&\sigma = \beta - \gamma - \theta + \delta,\\
&&W=(\gamma -\delta)P(y_x|c)+\delta P(y_{x'}|c)+\theta P(y'_{x'}|c),\\
&&L=\max\left\{
\begin{array}{c}
0,\\
P(y_x|c)-P(y_{x'}|c),\\
P(y|c)-P(y_{x'}|c),\\
P(y_x|c)-P(y|c)\\
\end{array}
\right\},\\
&&U=\min\left\{
\begin{array}{c}
P(y_x|c),\\
P(y'_{x'}|c),\\
P(y,x|c)+P(y',x'|c),\\
P(y_x|c)-P(y_{x'}|c)+\\+P(y,x'|c)+P(y',x|c)
\end{array}
\right\}.
\end{eqnarray*}
\end{theorem}
The above bounds of the benefit function used a combination of experimental and observational data. If the observational data is unavailable, then the bounds of the benefit function become
\begin{theorem}
\label{thm4}
Given a causal diagram $G$ and distribution compatible with $G$, let $C$ be a set of variables that does not contain any descendant of $X$ in $G$, then the benefit function $f(c)=\beta P(y_x,y'_{x'}|c)+\gamma P(y_x,y_{x'}|c)+ \theta P(y'_x,y'_{x'}|c) + \delta P(y_{x'},y'_{x}|c)$ is bounded as follows \cite{li:21-r507}:
\begin{eqnarray*}
W+\sigma U\le f(c) \le W+\sigma L\text{~~~~~~~~if }\sigma < 0,\\
W+\sigma L\le f(c) \le W+\sigma U\text{~~~~~~~~if }\sigma > 0,
\end{eqnarray*}
where $\sigma, W,L,U$ are given by,
\begin{eqnarray*}
&&\sigma = \beta - \gamma - \theta + \delta,\\
&&W=(\gamma -\delta)P(y_x|c)+\delta P(y_{x'}|c)+\theta P(y'_{x'}|c),\\
&&L=\max\left\{
\begin{array}{c}
0,\\
P(y_x|c)-P(y_{x'}|c)
\end{array}
\right\},\\
&&U=\min\left\{
\begin{array}{c}
P(y_x|c),\\
P(y'_{x'}|c)
\end{array}
\right\}.
\end{eqnarray*}
\end{theorem}

\subsection{Role of Observational Data in Unit Selection Problem}
The benefit function is a linear combination of probabilities of causation; thus, our Theorems \ref{thm1} and \ref{thm2} could be easily extended to the benefit function as follows:

\begin{corollary}
Given a causal diagram $G$ and distribution compatible with $G$ with experimental data $P(y_x|c)$ and $P(y_{x'}|c)$, let $C$ be a set of variables that does not contain any descendant of $X$ in $G$. let $D=P(y|c)$ be a random variable, and let $D'=P(x,y|c)+P(x',y'|c)$ be another random variable. Let $LB,UB$ be the lower and upper bound of the benefit function using pure experimental data, respectively. Let $LB',UB'$ be the lower and upper bound of the benefit function using a combination of experimental and observational data, respectively. If $D$ is uniformly distributed on its feasible interval $[max(0, P(y_x|c)-P({y'}_{x'}|c)), min(1, P(y_x|c)+P(y_{x'}|c))]$ and $D'$ is uniformly distributed on its feasible interval $[max(0, P(y_x|c)-P(y_{x'}|c)), min(1, P(y_x)+ P({y'}_{x'}|c))]$, and $P(y_x|c)+P({y'}_{x'}|c)\ne 0$ and $P({y'}_x|c)+P({y}_{x'}|c)\ne 0$ and $P(y_x|c)+P(y_{x'}|c)\ne 0$ and $P({y'}_x|c)+P({y'}_{x'}|c)\ne 0$, then we have the expectation of the increased lower bound $E(LB'-LB)$ and the decreased upper bound $E(UB-UB')$ as follows:
\begin{eqnarray*}
&&E(LB'-LB)\\
&=&\sigma\frac{\min\{P^2(y_x|c), P^2({y'}_x|c), P^2({y}_{x'}|c), P^2({y'}_{x'}|c)\}}{\min\{P(y_x|c)+P(y_{x'}|c),P({y'}_x|c)+P({y'}_{x'}|c)\}}\\
&&\text{~~~~~if }\sigma > 0,\\
&&E(LB'-LB)\\
&=&-\sigma\frac{\min\{P^2(y_x|c), P^2({y'}_x|c), P^2({y}_{x'}|c), P^2({y'}_{x'}|c)\}}{\min\{P(y_x|c)+P({y'}_{x'}|c),P({y'}_x|c)+P({y}_{x'}|c)\}}\\
&&\text{~~~~~if }\sigma < 0,\\
&&E(UB-UB')\\
&=&\sigma\frac{\min\{P^2(y_x|c), P^2({y'}_x|c), P^2({y}_{x'}|c), P^2({y'}_{x'}|c)\}}{\min\{P(y_x|c)+P({y'}_{x'}|c),P({y'}_x|c)+P({y}_{x'}|c)\}}\\
&&\text{~~~~~if }\sigma > 0,\\
&&E(UB-UB')\\
&=&-\sigma\frac{\min\{P^2(y_x|c), P^2({y'}_x|c), P^2({y}_{x'}|c), P^2({y'}_{x'}|c)\}}{\min\{P(y_x|c)+P(y_{x'}|c),P({y'}_x|c)+P({y'}_{x'}|c)\}}\\
&&\text{~~~~~if }\sigma < 0,
\end{eqnarray*}
where,
\begin{eqnarray*}
&&\text{the benefit function~} f(c)=\beta P(y_x,y'_{x'}|c)+\\
&&+\gamma P(y_x,y_{x'}|c)+\theta P(y'_x,y'_{x'}|c) + \delta P(y_{x'},y'_{x}|c),\\
&&\sigma = \beta - \gamma - \theta + \delta,\\
&&W=(\gamma -\delta)P(y_x|c)+\delta P(y_{x'}|c)+\theta P(y'_{x'}|c),\\
&&P({y'}_x|c) = 1 - P(y_x|c),\\
&&P({y'}_{x'}|c) = 1 - P(y_{x'}|c),\\
&&LB = W+\sigma U\text{~~~~~~~~if }\sigma < 0,\\
&&LB = W+\sigma L\text{~~~~~~~~if }\sigma > 0,\\
&&LB' = W+\sigma U'\text{~~~~~~~~if }\sigma < 0,\\
&&LB' = W+\sigma L'\text{~~~~~~~~if }\sigma > 0,\\
&&UB = W+\sigma L\text{~~~~~~~~if }\sigma < 0,\\
&&UB = W+\sigma U\text{~~~~~~~~if }\sigma > 0,\\
&&UB' = W+\sigma L'\text{~~~~~~~~if }\sigma < 0,\\
&&UB' = W+\sigma U'\text{~~~~~~~~if }\sigma > 0,\\
&&\text{L'} = \max \left \{
\begin{array}{cc}
0, \\
P(y_x|c) - P(y_{x'}|c), \\
D - P(y_{x'}|c), \\
P(y_x|c) - D
\end{array}
\right \},\\
&&\text{L} = \max \left \{
\begin{array}{cc}
0, \\
P(y_x|c) - P(y_{x'}|c)
\end{array}
\right \},\\
&&\text{U'} = \min \left \{
\begin{array}{cc}
 P(y_x|c), \\
 P(y'_{x'}|c), \\
D', \\
P(y_x|c) + P({y'}_{x'}|c) -D'
\end{array}
\right \},\\
&&\text{U} = \min \left \{
\begin{array}{cc}
 P(y_x|c), \\
 P(y'_{x'}|c)
\end{array}
\right \}.
\end{eqnarray*}
\label{thm5}
\end{corollary}
Note that $\sigma = 0$ is the case called Gain Equality \cite{li:pea19-r488} where the benefit function is reduced to a point estimation with purely experimental data. The $\sigma$ here can be interpreted as the improvement amplifier. This corollary looks more complicated than Theorems \ref{thm1} and \ref{thm2}, but the usage is exactly the same as the Theorems \ref{thm1} and \ref{thm2}. We will illustrate how to apply this corollary in the next section.
\subsection{Nonimmediate Profit}
Consider a car dealer again who wants to send an enticement (i.e., discount) to all customers to increase its total nonimmediate profit. The manager acknowledged that the payoff of sending the discount to a complier (i.e., the customer who would buy the hybrid car if they received the discount and would not otherwise) is $\$1500$ as the profit of selling a hybrid car is $\$2000$, but the discount is $\$500$. The payoff of sending the discount to an always-taker (i.e., the customer who would buy the hybrid car no matter whether or not they received the discount) is $-\$800$. First, the dealer loses the cost of the discount, $\$500$. Second, the customer may require a discount in the future (i.e., the offer changed the customer's response type, and the manager assessed that this nonimmediate cost is $\$300$). The payoff of sending the discount to a never-taker (i.e., the customer who would not buy the hybrid car no matter whether or not they received the discount) is $\$0$, as the cost of sending the discount is negligible. The payoff of sending the discount to a defier (i.e., the customer who would buy the hybrid car if they received no discount and would not otherwise) is $-\$2000$ as the car dealer loses one customer due to the discount.

Again, let $X=x$ denote that a customer received the discount and $X=x'$ denote that a customer received no discount. Let $Y=y$ denote that a customer purchased the hybrid car and $Y=y'$ denote that a customer did not purchase the hybrid car. Note that the population-specific characteristics $c$ here simply mean the entire customer.

The benefit vector is then $(1500, -800, 0, -2000)$. The manager conducted an experimental study with $3000$ randomly selected customers. They sent the discount to $1500$ of the selected customers and sent no discount to the rest $1500$ of the selected customers. The results are shown in Table \ref{tb4}.

\begin{table}
\centering
\caption{Results of an experimental study where $1500$ customers were forced to receive the discount and $1500$ customers were forced to not.}
\begin{tabular}{|c|c|c|}
\hline
&Received discount&No discount\\
\hline
Purchased&$825$&$600$\\
\hline
Not purchased&$675$&$900$\\
\hline
\end{tabular}
\label{tb4}
\end{table}

The experimental data are then $P(y_x|c)=0.55$ and $P(y_{x'}|c)=0.4$. We then obtained that $-95 \le f(c)\le 20$ by Theorem \ref{thm4} using the experimental data. These results are hard for the decision-maker because the average gain profit of sending the discount to a customer is across the $0$ (i.e., either gain profit or lose profit).

An observational study was then considered by the manager; therefore, the manager applied the Corollary \ref{thm5} and obtained that $E(LB'-LB)=50.53$ and $E(UB-UB')=56.47$, where the expected new upper bound of the benefit function may smaller than $0$.

The manager then conducted an observational study where $1500$ randomly selected customers had access to the discount and were free to choose to receive it or not. $795$ of the selected customers chose to receive the discount, and $705$ of the selected customers decided not to. The results are shown in Table \ref{tb5}.

\begin{table}
\centering
\caption{Results of an observational study with $1500$ customers who have access to the discount, where $795$ customers chose to receive the discount and $705$ customers chose not to.}
\begin{tabular}{|c|c|c|}
\hline
&Chose discount&No discount\\
\hline
Purchased&$450$&$30$\\
\hline
Not purchased&$345$&$675$\\
\hline
\end{tabular}
\label{tb5}
\end{table}

The observational data are then $P(x,y)=0.3$, $P(x',y)=0.02$, $P(x,y')=0.23$, and $P(x',y')=0.45$. The bounds of the benefit function obtained from Theorem \ref{thm3} is then $-71 \le f(c)\le -20$, where the upper bound is smaller than $0$. The conclusion is then not to send the discount because the average profit gained per customer by sending the discount is negative (i.e., the car dealer will lose the profit if sending the discount to all customers).

\section{DISCUSSION}
The paper provides theorems that quantify the degree of improvement one can expect from observational data, when considering the bounds of PNS. Here, we further discuss two additional topics related to the theorems provided.

First, while this paper answers the question of whether the observational studies are worth conducting, the symmetrical problem can also be
answered. If experimental data are unavailable, the bounds of PNS are $0\le PNS \le P(x,y)+(x',y')$ which are less informative as discussed in \cite{li:21-r507}. In such cases, experimental data should be investigated either by random controlled trial or by adjustment formula \cite{pearl1993aspects}.

Second, the assumptions in Theorems \ref{thm1} and \ref{thm2} are $P(y)$ and $P(x,y)+P(x',y')$ are uniformly distributed on their feasible interval. We know that given large enough observational samples of $(X,Y)$, the sample estimation of a fixed $P(y)$ (or $P(x,y)+P(x',y')$) is normally distributed with the center at $P(y)$ (or $P(x,y)+P(x',y')$). However, the distribution of $P(y)$ (or $P(x,y)+P(x',y')$) is unknown; one can interpret that this distribution is among all compatible models in the world (e.g., $P(y)=0.1$ in one model and $P(y)=0.5$ in another model). If there is any progress on these distributions, the uniform distribution assumptions in both theorems can be extended.
\section{CONCLUSION}
In this work, we formally defined the expectation of improvement of the bounds of PNS when considering observational data. The proposed theorems are extendable to other probabilities of causation. Simulated examples showed that the proposed theorems should be applied before conducting any observational study. We further expanded the proposed theorems to the bounds of the benefit function in the unit selection problem.

\subsubsection*{Acknowledgements}
This research was supported in parts by grants from the National Science
Foundation [\#IIS-2106908 and \#IIS-2231798], Office of Naval Research [\#N00014-21-1-2351], and Toyota Research Institute of North America
[\#PO-000897].

\bibliographystyle{plain}
\bibliography{aistats2023.bib}
\clearpage
\newpage
\section{APPENDIX}
\subsection{Proof of Theorem \ref{thm1}}
\begin{reptheorem}{thm1}
Given experimental data $P(y_x)$ and $P(y_{x'})$ and let $D=P(y)$ be a random variable. Let $L$ be the lower bound of PNS using pure experimental data and $L'$ be the lower bound of PNS using a combination of experimental and observational data. If $D$ is uniformly distributed on its feasible interval $[max(0, P(y_x)-P({y'}_{x'})), min(1, P(y_x)+P(y_{x'}))]$ ,and $P(y_x)+P(y_{x'})\ne 0$ and $P({y'}_x)+P({y'}_{x'})\ne 0$, then we have the expectation of the increased lower bound $E(L'-L)$ as follows:
\begin{eqnarray*}
&&E(L'-L)\\
&=&\frac{\min\{P^2(y_x), P^2({y'}_x), P^2({y}_{x'}), P^2({y'}_{x'})\}}{\min\{P(y_x)+P(y_{x'}),P({y'}_x)+P({y'}_{x'})\}}
\end{eqnarray*}
where,\\
\begin{eqnarray*}
P({y'}_x) = 1 - P(y_x),\\
P({y'}_{x'}) = 1 - P(y_{x'}),
\end{eqnarray*}
\begin{eqnarray*}
\text{L'} = \max \left \{
\begin{array}{cc}
0, \\
P(y_x) - P(y_{x'}), \\
D - P(y_{x'}), \\
P(y_x) - D
\end{array}
\right \},
\end{eqnarray*}
\begin{eqnarray*}
\text{L} = \max \left \{
\begin{array}{cc}
0, \\
P(y_x) - P(y_{x'})
\end{array}
\right \}.
\end{eqnarray*}
\begin{proof}
First, by Tian and Pearl \cite{tian2000probabilities}, we have,
\begin{eqnarray*}
P(x,y)\le P(y_x)\le 1 - P(x,y'),\\
P(x',y)\le P(y_{x'})\le 1 - P(x',y').\\
\end{eqnarray*}
Therefore,
\begin{eqnarray*}
P(y_x)+P(y_{x'}) &\ge& P(x,y) + P(x',y)\\
&=&P(y),
\end{eqnarray*}
and,
\begin{eqnarray*}
P(y_x)+P(y_{x'}) &\le& 2 - P(x,y') - P(x',y')\\
&=&2 - P(y')\\
&=& 1 + P(y),\\
P(y) &\ge & P(y_x)+P(y_{x'}) - 1\\
&=& P(y_x)-P({y'}_{x'}).
\end{eqnarray*}
Thus, the feasible space of $D=P(y)$ is
\begin{eqnarray*}
D&\ge& \max\{0, P(y_x)-P({y'}_{x'})\},\\
D&\le& \min\{1, P(y_x)+P(y_{x'})\}.
\end{eqnarray*}
Case 1: $P(y_x) \ge P(y_{x'})$ and $P(y_x)+P(y_{x'})\ge 1$.\\
Thus,
\begin{eqnarray*}
P(y_x)+P(y_{x'}) - 1&\ge& 0,\\
P(y_x)-P({y'}_{x'}) &\ge& 0.
\end{eqnarray*}
Therefore, the feasible space of $D$ is 
\begin{eqnarray*}
P(y_x)-P({y'}_{x'}) \le D \le 1.
\end{eqnarray*}
and $D$ uniformly distributed on $[P(y_x)-P({y'}_{x'}), 1]$.\\
Also, $L=\max\{0, P(y_x) - P(y_{x'})\}=P(y_x) - P(y_{x'})$,\\
and $L' \ge P(y_x) - P(y_{x'})$.
In order to make $L'>L$, $D$ has to be in interval $[P(y_x)-P({y'}_{x'}), P(y_{x'}))$ or $(P(y_x), 1]$.\\
We also know that,
\begin{eqnarray*}
1 - P(y_x) = P(y_{x'}) - P(y_x) + P({y'}_{x'}) = P({y'}_x),
\end{eqnarray*}
therefore, the maximum value of $L'-L$ is $P({y'}_x)$,\\
and $L'-L \le z \le P({y'}_x)$, when $D$ in interval $[P(y_{x'}) - z, P(y_{x'})]$ or $[P(y_x), P(y_x) + z]$.\\
Thus, we have,
\begin{eqnarray*}
&&P(L'-L \le z)\\
&=&\left \{
\begin{array}{lr}
0, \text{ ~~~~~~~~~~~~~~~~~~~~~~~~~if } z < 0,\\
\frac{P(y_x) - P(y_{x'}) +2z}{P({y'}_x) + P({y'}_{x'})}, \text{ if }  0 \le z < P({y'}_x),\\
1, \text{ ~~~~~~~~~~~~~~~~~~~~~~~~~if } z \ge P({y'}_x).
\end{array}
\right \},
\end{eqnarray*}
Then, the probability density function $f$ is\\
\begin{eqnarray*}
&&f(z)\\
&=&\left \{
\begin{array}{lr}
0, \text{ ~~~~~~~~~~~~~~~~~~~~~if } z < 0,\\
\frac{2}{P({y'}_x) + P({y'}_{x'})}, \text{ if }  0 \le z < P({y'}_x),\\
0, \text{ ~~~~~~~~~~~~~~~~~~~~~if } z \ge P({y'}_x).
\end{array}
\right \},
\end{eqnarray*}
Then, we have,
\begin{eqnarray*}
&&E(L'-L)\\
&=&\int_{0}^{P({y'}_x)}\frac{2z}{P({y'}_x) + P({y'}_{x'})}dz\\
&=&\frac{P^2({y'}_x)}{P({y'}_x) + P({y'}_{x'})}.
\end{eqnarray*}

Case 2: $P(y_x) \ge P(y_{x'})$ and $P(y_x)+P(y_{x'})< 1$.\\
Thus,
\begin{eqnarray*}
P(y_x)+P(y_{x'}) - 1&<& 0,\\
P(y_x)-P({y'}_{x'}) &<& 0.
\end{eqnarray*}
Therefore, the feasible space of $D$ is 
\begin{eqnarray*}
0 \le D \le P(y_x)+P(y_{x'}).
\end{eqnarray*}
and $D$ uniformly distributed on $[0, P(y_x)+P(y_{x'})]$.\\
Also, $L=\max\{0, P(y_x) - P(y_{x'})\}=P(y_x) - P(y_{x'})$,\\
and $L' \ge P(y_x) - P(y_{x'})$.
In order to make $L'>L$, $D$ has to be in interval $[0, P(y_{x'}))$ or $(P(y_x), P(y_x)+P(y_{x'})]$.\\
We also know that,
\begin{eqnarray*}
P(y_{x'}) - 0 = P(y_x)+P(y_{x'}) - P(y_x) = P(y_{x'}),
\end{eqnarray*}
therefore, the maximum value of $L'-L$ is $P({y}_{x'})$,\\
and $L'-L \le z \le P({y}_{x'})$, when $D$ in interval $[P(y_{x'}) - z, P(y_{x'})]$ or $[P(y_x), P(y_x) + z]$.\\
Thus, we have,
\begin{eqnarray*}
&&P(L'-L \le z)\\
&=&\left \{
\begin{array}{lr}
0, \text{ ~~~~~~~~~~~~~~~~~~~~~~~~~if } z < 0,\\
\frac{P(y_x) - P(y_{x'}) +2z}{P(y_x)+P(y_{x'})}, \text{ if }  0 \le z < P({y}_{x'}),\\
1, \text{ ~~~~~~~~~~~~~~~~~~~~~~~~~if } z \ge P({y}_{x'}).
\end{array}
\right \},
\end{eqnarray*}
Then, the probability density function $f$ is\\
\begin{eqnarray*}
&&f(z)\\
&=&\left \{
\begin{array}{lr}
0, \text{ ~~~~~~~~~~~~~~~~~~~~~if } z < 0,\\
\frac{2}{P(y_x)+P(y_{x'})}, \text{ if }  0 \le z < P({y}_{x'}),\\
0, \text{ ~~~~~~~~~~~~~~~~~~~~~if } z \ge P({y}_{x'}).
\end{array}
\right \},
\end{eqnarray*}
Then, we have,
\begin{eqnarray*}
&&E(L'-L)\\
&=&\int_{0}^{P({y}_{x'})}\frac{2z}{P(y_x)+P(y_{x'})}dz\\
&=&\frac{P^2({y}_{x'})}{P(y_x)+P(y_{x'})}.
\end{eqnarray*}

Case 3: $P(y_x) < P(y_{x'})$ and $P(y_x)+P(y_{x'})\ge 1$.\\
Thus,
\begin{eqnarray*}
P(y_x)+P(y_{x'}) - 1&\ge& 0,\\
P(y_x)-P({y'}_{x'}) &\ge& 0.
\end{eqnarray*}
Therefore, the feasible space of $D$ is 
\begin{eqnarray*}
P(y_x)-P({y'}_{x'}) \le D \le 1.
\end{eqnarray*}
and $D$ uniformly distributed on $[P(y_x)-P({y'}_{x'}), 1]$.\\
Also, $L=\max\{0, P(y_x) - P(y_{x'})\}=0$,\\
and $L' \ge 0$.
In order to make $L'>L$, $D$ has to be in interval $[P(y_x)-P({y'}_{x'}), P(y_{x}))$ or $(P(y_{x'}), 1]$.\\
We also know that,
\begin{eqnarray*}
1 - P(y_{x'}) = P(y_{x}) - P(y_x) + P({y'}_{x'}) = P({y'}_{x'}),
\end{eqnarray*}
therefore, the maximum value of $L'-L$ is $P({y'}_{x'})$,\\
and $L'-L \le z \le P({y'}_{x'})$, when $D$ in interval $[P(y_{x}) - z, P(y_{x})]$ or $[P(y_{x'}), P(y_{x'}) + z]$.\\
Thus, we have,
\begin{eqnarray*}
&&P(L'-L \le z)\\
&=&\left \{
\begin{array}{lr}
0, \text{ ~~~~~~~~~~~~~~~~~~~~~~~~~if } z < 0,\\
\frac{P(y_{x'}) - P(y_{x}) +2z}{P({y'}_x) + P({y'}_{x'})}, \text{ if }  0 \le z < P({y'}_{x'}),\\
1, \text{ ~~~~~~~~~~~~~~~~~~~~~~~~~if } z \ge P({y'}_{x'}).
\end{array}
\right \},
\end{eqnarray*}
Then, the probability density function $f$ is\\
\begin{eqnarray*}
&&f(z)\\
&=&\left \{
\begin{array}{lr}
0, \text{ ~~~~~~~~~~~~~~~~~~~~~if } z < 0,\\
\frac{2}{P({y'}_x) + P({y'}_{x'})}, \text{ if }  0 \le z < P({y'}_{x'}),\\
0, \text{ ~~~~~~~~~~~~~~~~~~~~~if } z \ge P({y'}_{x'}).
\end{array}
\right \},
\end{eqnarray*}
Then, we have,
\begin{eqnarray*}
&&E(L'-L)\\
&=&\int_{0}^{P({y'}_{x'})}\frac{2z}{P({y'}_x) + P({y'}_{x'})}dz\\
&=&\frac{P^2({y'}_{x'})}{P({y'}_x) + P({y'}_{x'})}.
\end{eqnarray*}

Case 4: $P(y_x) < P(y_{x'})$ and $P(y_x)+P(y_{x'})< 1$.\\
Thus,
\begin{eqnarray*}
P(y_x)+P(y_{x'}) - 1&<& 0,\\
P(y_x)-P({y'}_{x'}) &<& 0.
\end{eqnarray*}
Therefore, the feasible space of $D$ is 
\begin{eqnarray*}
0 \le D \le P(y_x)+P(y_{x'}).
\end{eqnarray*}
and $D$ uniformly distributed on $[0, P(y_x)+P(y_{x'})]$.\\
Also, $L=\max\{0, P(y_x) - P(y_{x'})\}=0$,\\
and $L' \ge 0$.
In order to make $L'>L$, $D$ has to be in interval $[0, P(y_{x}))$ or $(P(y_{x'}), P(y_x)+P(y_{x'})]$.\\
We also know that,
\begin{eqnarray*}
P(y_{x}) - 0 = P(y_x)+P(y_{x'}) - P(y_{x'}) = P(y_{x}),
\end{eqnarray*}
therefore, the maximum value of $L'-L$ is $P({y}_{x})$,\\
and $L'-L \le z \le P({y}_{x})$, when $D$ in interval $[P(y_{x}) - z, P(y_{x})]$ or $[P(y_{x'}), P(y_{x'}) + z]$.\\
Thus, we have,
\begin{eqnarray*}
&&P(L'-L \le z)\\
&=&\left \{
\begin{array}{lr}
0, \text{ ~~~~~~~~~~~~~~~~~~~~~~~~~if } z < 0,\\
\frac{P(y_{x'}) - P(y_{x}) +2z}{P(y_x)+P(y_{x'})}, \text{ if }  0 \le z < P({y}_{x}),\\
1, \text{ ~~~~~~~~~~~~~~~~~~~~~~~~~if } z \ge P({y}_{x}).
\end{array}
\right \},
\end{eqnarray*}
Then, the probability density function $f$ is\\
\begin{eqnarray*}
&&f(z)\\
&=&\left \{
\begin{array}{lr}
0, \text{ ~~~~~~~~~~~~~~~~~~~~~if } z < 0,\\
\frac{2}{P(y_x)+P(y_{x'})}, \text{ if }  0 \le z < P({y}_{x}),\\
0, \text{ ~~~~~~~~~~~~~~~~~~~~~if } z \ge P({y}_{x}).
\end{array}
\right \},
\end{eqnarray*}
Then, we have,
\begin{eqnarray*}
&&E(L'-L)\\
&=&\int_{0}^{P({y}_{x})}\frac{2z}{P(y_x)+P(y_{x'})}dz\\
&=&\frac{P^2({y}_{x})}{P(y_x)+P(y_{x'})}.
\end{eqnarray*}
We then combine the results of cases $1$ and $3$.\\
Note that when $P(y_x)\ge P(y_{x'})$, we have $P({y'}_x)\le P({y'}_{x'})$,\\
and when $P(y_x) < P(y_{x'})$, we have $P({y'}_x)> P({y'}_{x'})$,\\
therefore,\\
when $P(y_x)+P(y_{x'})\ge 1$, we have,
\begin{eqnarray*}
&&E(L'-L)\\
&=&\frac{\min\{P^2({y'}_{x}), P^2({y'}_{x'})\}}{P({y'}_x) + P({y'}_{x'})}.
\end{eqnarray*}
Similarly, after combine the results of cases $2$ and $4$, we have,\\
when $P(y_x)+P(y_{x'}) <  1$, we have,
\begin{eqnarray*}
&&E(L'-L)\\
&=&\frac{\min\{P^2({y}_{x}), P^2({y}_{x'})\}}{P({y}_x) + P({y}_{x'})}.
\end{eqnarray*}
Now, note that when $P(y_x)+P(y_{x'})\ge 1$, we have,\\
$P(y_x)+P(y_{x'})\ge P({y'}_x)+P({y'}_{x'})$.\\
Also, $P(y_x)\ge P({y'}_{x'})$ and $P(y_{x'})\ge P({y'}_{x})$,\\
thus, $\min\{P^2({y}_{x}), P^2({y}_{x'})\} \ge \min\{P^2({y'}_{x}), P^2({y'}_{x'})\}$.\\
Similarly, when $P(y_x)+P(y_{x'})< 1$, we have,\\
$P(y_x)+P(y_{x'}) < P({y'}_x)+P({y'}_{x'})$,\\
and, $\min\{P^2({y}_{x}), P^2({y}_{x'})\} < \min\{P^2({y'}_{x}), P^2({y'}_{x'})\}$.
Thus, finally, we have,
\begin{eqnarray*}
&&E(L'-L)\\
&=&\frac{\min\{P^2(y_x), P^2({y'}_x), P^2({y}_{x'}), P^2({y'}_{x'})\}}{\min\{P(y_x)+P(y_{x'}),P({y'}_x)+P({y'}_{x'})\}}.
\end{eqnarray*}
\end{proof}
\end{reptheorem}

\subsection{Proof of Theorem \ref{thm2}}
\begin{reptheorem}{thm2}
Given experimental data $P(y_x)$ and $P(y_{x'})$ and let $D=P(x,y)+P(x',y')$ be a random variable. Let $U$ be the upper bound of PNS using pure experimental data and $U'$ be the upper bound of PNS using a combination of experimental and observational data. If $D$ is uniformly distributed on its feasible interval $[max(0, P(y_x)-P(y_{x'})), min(1, P(y_x)+ P({y'}_{x'}))]$, and $P(y_x)+P({y'}_{x'})\ne 0$ and $P({y'}_x)+P({y}_{x'})\ne 0$, then we have the expectation of the decreased upper bound $E(U-U')$ as follows:
\begin{eqnarray*}
&&E(U-U')\\
&=&\frac{\min\{P^2(y_x), P^2({y'}_x), P^2({y}_{x'}), P^2({y'}_{x'})\}}{\min\{P(y_x)+P({y'}_{x'}),P({y'}_x)+P({y}_{x'})\}}
\end{eqnarray*}
where,
\begin{eqnarray*}
P({y'}_x) = 1 - P(y_x),\\
P({y'}_{x'}) = 1 - P(y_{x'}),
\end{eqnarray*}
\begin{eqnarray*}
\text{U'} = \min \left \{
\begin{array}{cc}
 P(y_x), \\
 P(y'_{x'}), \\
D, \\
P(y_x) + P({y'}_{x'}) -D
\end{array}
\right \},
\end{eqnarray*}
\begin{eqnarray*}
\text{U} = \min \left \{
\begin{array}{cc}
 P(y_x), \\
 P(y'_{x'})
\end{array}
\right \}.
\end{eqnarray*}
\begin{proof}
First, by Tian and Pearl \cite{tian2000probabilities}, we have,
\begin{eqnarray*}
P(x,y)\le P(y_x)\le 1 - P(x,y'),\\
P(x',y')\le P({y'}_{x'})\le 1 - P(x',y).\\
\end{eqnarray*}
Therefore,
\begin{eqnarray*}
P(y_x)+P({y'}_{x'}) &\ge& P(x,y) + P(x',y'),
\end{eqnarray*}
and,
\begin{eqnarray*}
P(y_x)+P({y'}_{x'}) &\le& 2 - P(x,y') - P(x',y)\\
&=& 1 + P(x,y)+P(x',y'),\\
P(x,y) +P(x',y') &\ge & P(y_x)+P({y'}_{x'}) - 1\\
&=& P(y_x)-P({y}_{x'}).
\end{eqnarray*}
Thus, the feasible space of $D=P(x,y)+P(x',y')$ is
\begin{eqnarray*}
D&\ge& \max\{0, P(y_x)-P({y}_{x'})\},\\
D&\le& \min\{1, P(y_x)+P({y'}_{x'})\}.
\end{eqnarray*}
Case 1: $P(y_x) \ge P({y'}_{x'})$ and $P(y_x)+P({y'}_{x'})\ge 1$.\\
Thus,
\begin{eqnarray*}
P(y_x)+P({y'}_{x'}) - 1&\ge& 0,\\
P(y_x)-P({y}_{x'}) &\ge& 0.
\end{eqnarray*}
Therefore, the feasible space of $D$ is 
\begin{eqnarray*}
P(y_x)-P({y}_{x'}) \le D \le 1.
\end{eqnarray*}
and $D$ uniformly distributed on $[P(y_x)-P({y}_{x'}), 1]$.\\
Also, $U=\min\{P(y_x), P({y'}_{x'})\}=P({y'}_{x'})$,\\
and $U' \le P({y'}_{x'})$.
In order to make $U'<U$, $D$ has to be in interval $[P(y_x)-P({y}_{x'}), P({y'}_{x'}))$ or $(P(y_x), 1]$.\\
We also know that,
\begin{eqnarray*}
1 - P(y_x) = P({y'}_{x'}) - P(y_x) + P({y}_{x'}) = P({y'}_x),
\end{eqnarray*}
therefore, the maximum value of $U-U'$ is $P({y'}_x)$,\\
and $U-U' \le z \le P({y'}_x)$, when $D$ in interval $[P({y'}_{x'}) - z, P({y'}_{x'})]$ or $[P(y_x), P(y_x) + z]$.\\
Thus, we have,
\begin{eqnarray*}
&&P(U-U' \le z)\\
&=&\left \{
\begin{array}{lr}
0, \text{ ~~~~~~~~~~~~~~~~~~~~~~~~~if } z < 0,\\
\frac{P(y_x) - P({y'}_{x'}) +2z}{P({y'}_x) + P({y}_{x'})}, \text{ if }  0 \le z < P({y'}_x),\\
1, \text{ ~~~~~~~~~~~~~~~~~~~~~~~~~if } z \ge P({y'}_x).
\end{array}
\right \},
\end{eqnarray*}
Then, the probability density function $f$ is\\
\begin{eqnarray*}
&&f(z)\\
&=&\left \{
\begin{array}{lr}
0, \text{ ~~~~~~~~~~~~~~~~~~~~~if } z < 0,\\
\frac{2}{P({y'}_x) + P({y}_{x'})}, \text{ if }  0 \le z < P({y'}_x),\\
0, \text{ ~~~~~~~~~~~~~~~~~~~~~if } z \ge P({y'}_x).
\end{array}
\right \},
\end{eqnarray*}
Then, we have,
\begin{eqnarray*}
&&E(U-U')\\
&=&\int_{0}^{P({y'}_x)}\frac{2z}{P({y'}_x) + P({y}_{x'})}dz\\
&=&\frac{P^2({y'}_x)}{P({y'}_x) + P({y}_{x'})}.
\end{eqnarray*}

Case 2: $P(y_x) \ge P({y'}_{x'})$ and $P(y_x)+P({y'}_{x'})< 1$.\\
Thus,
\begin{eqnarray*}
P(y_x)+P({y'}_{x'}) - 1&<& 0,\\
P(y_x)-P({y}_{x'}) &<& 0.
\end{eqnarray*}
Therefore, the feasible space of $D$ is 
\begin{eqnarray*}
0 \le D \le P(y_x)+P({y'}_{x'}).
\end{eqnarray*}
and $D$ uniformly distributed on $[0, P(y_x)+P({y'}_{x'})]$.\\
Also, $U=\min\{P(y_x),P({y'}_{x'})\}=P({y'}_{x'})$,\\
and $U' < P({y'}_{x'})$.
In order to make $U'<U$, $D$ has to be in interval $[0, P({y'}_{x'}))$ or $(P(y_x), P(y_x)+P({y'}_{x'})]$.\\
We also know that,
\begin{eqnarray*}
P({y'}_{x'}) - 0 = P(y_x)+P({y'}_{x'}) - P(y_x) = P({y'}_{x'}),
\end{eqnarray*}
therefore, the maximum value of $U-U'$ is $P({y'}_{x'})$,\\
and $U-U' \le z \le P({y'}_{x'})$, when $D$ in interval $[P({y'}_{x'}) - z, P({y'}_{x'})]$ or $[P(y_x), P(y_x) + z]$.\\
Thus, we have,
\begin{eqnarray*}
&&P(U-U' \le z)\\
&=&\left \{
\begin{array}{lr}
0, \text{ ~~~~~~~~~~~~~~~~~~~~~~~~~if } z < 0,\\
\frac{P(y_x) - P({y'}_{x'}) +2z}{P(y_x)+P({y'}_{x'})}, \text{ if }  0 \le z < P({y'}_{x'}),\\
1, \text{ ~~~~~~~~~~~~~~~~~~~~~~~~~if } z \ge P({y'}_{x'}).
\end{array}
\right \},
\end{eqnarray*}
Then, the probability density function $f$ is\\
\begin{eqnarray*}
&&f(z)\\
&=&\left \{
\begin{array}{lr}
0, \text{ ~~~~~~~~~~~~~~~~~~~~~if } z < 0,\\
\frac{2}{P(y_x)+P({y'}_{x'})}, \text{ if }  0 \le z < P({y'}_{x'}),\\
0, \text{ ~~~~~~~~~~~~~~~~~~~~~if } z \ge P({y'}_{x'}).
\end{array}
\right \},
\end{eqnarray*}
Then, we have,
\begin{eqnarray*}
&&E(U-U')\\
&=&\int_{0}^{P({y'}_{x'})}\frac{2z}{P(y_x)+P({y'}_{x'})}dz\\
&=&\frac{P^2({y'}_{x'})}{P(y_x)+P({y'}_{x'})}.
\end{eqnarray*}

Case 3: $P(y_x) < P({y'}_{x'})$ and $P(y_x)+P({y'}_{x'})\ge 1$.\\
Thus,
\begin{eqnarray*}
P(y_x)+P({y'}_{x'}) - 1&\ge& 0,\\
P(y_x)-P({y}_{x'}) &\ge& 0.
\end{eqnarray*}
Therefore, the feasible space of $D$ is 
\begin{eqnarray*}
P(y_x)-P({y}_{x'}) \le D \le 1.
\end{eqnarray*}
and $D$ uniformly distributed on $[P(y_x)-P({y}_{x'}), 1]$.\\
Also, $U=\min\{P(y_x), P({y'}_{x'})\}=P({y}_{x})$,\\
and $U' \le P({y}_{x})$.
In order to make $U'<U$, $D$ has to be in interval $[P(y_x)-P({y}_{x'}), P({y}_{x}))$ or $(P({y'}_{x'}), 1]$.\\
We also know that,
\begin{eqnarray*}
1 - P({y'}_{x'}) = P({y}_{x}) - P(y_x) + P({y}_{x'}) = P({y}_{x'}),
\end{eqnarray*}
therefore, the maximum value of $U-U'$ is $P({y}_{x'})$,\\
and $U-U' \le z \le P({y}_{x'})$, when $D$ in interval $[P({y}_{x}) - z, P({y}_{x})]$ or $[P({y'}_{x'}), P({y'}_{x'}) + z]$.\\
Thus, we have,
\begin{eqnarray*}
&&P(U-U' \le z)\\
&=&\left \{
\begin{array}{lr}
0, \text{ ~~~~~~~~~~~~~~~~~~~~~~~~~if } z < 0,\\
\frac{P({y'}_{x'}) - P({y}_{x}) +2z}{P({y'}_x) + P({y}_{x'})}, \text{ if }  0 \le z < P({y}_{x'}),\\
1, \text{ ~~~~~~~~~~~~~~~~~~~~~~~~~if } z \ge P({y}_{x'}).
\end{array}
\right \},
\end{eqnarray*}
Then, the probability density function $f$ is\\
\begin{eqnarray*}
&&f(z)\\
&=&\left \{
\begin{array}{lr}
0, \text{ ~~~~~~~~~~~~~~~~~~~~~if } z < 0,\\
\frac{2}{P({y'}_x) + P({y}_{x'})}, \text{ if }  0 \le z < P({y}_{x'}),\\
0, \text{ ~~~~~~~~~~~~~~~~~~~~~if } z \ge P({y}_{x'}).
\end{array}
\right \},
\end{eqnarray*}
Then, we have,
\begin{eqnarray*}
&&E(U-U')\\
&=&\int_{0}^{P({y}_{x'})}\frac{2z}{P({y'}_x) + P({y}_{x'})}dz\\
&=&\frac{P^2({y}_{x'})}{P({y'}_x) + P({y}_{x'})}.
\end{eqnarray*}

Case 4: $P(y_x) < P({y'}_{x'})$ and $P(y_x)+P({y'}_{x'})< 1$.\\
Thus,
\begin{eqnarray*}
P(y_x)+P({y'}_{x'}) - 1&<& 0,\\
P(y_x)-P({y}_{x'}) &<& 0.
\end{eqnarray*}
Therefore, the feasible space of $D$ is 
\begin{eqnarray*}
0 \le D \le P(y_x)+P({y'}_{x'}).
\end{eqnarray*}
and $D$ uniformly distributed on $[0, P(y_x)+P({y'}_{x'})]$.\\
Also, $U=\min\{P(y_x),P({y'}_{x'})\}=P({y}_{x})$,\\
and $U' < P({y}_{x})$.
In order to make $U'<U$, $D$ has to be in interval $[0, P({y}_{x}))$ or $(P({y'}_{x'}), P(y_x)+P({y'}_{x'})]$.\\
We also know that,
\begin{eqnarray*}
P({y}_{x}) - 0 = P(y_x)+P({y'}_{x'}) - P({y'}_{x'}) = P({y}_{x}),
\end{eqnarray*}
therefore, the maximum value of $U-U'$ is $P({y}_{x})$,\\
and $U-U' \le z \le P({y}_{x})$, when $D$ in interval $[P({y}_{x}) - z, P({y}_{x})]$ or $[P({y'}_{x'}), P({y'}_{x'}) + z]$.\\
Thus, we have,
\begin{eqnarray*}
&&P(U-U' \le z)\\
&=&\left \{
\begin{array}{lr}
0, \text{ ~~~~~~~~~~~~~~~~~~~~~~~~~if } z < 0,\\
\frac{P({y'}_{x'}) - P({y}_{x}) +2z}{P(y_x)+P({y'}_{x'})}, \text{ if }  0 \le z < P({y}_{x}),\\
1, \text{ ~~~~~~~~~~~~~~~~~~~~~~~~~if } z \ge P({y}_{x}).
\end{array}
\right \},
\end{eqnarray*}
Then, the probability density function $f$ is\\
\begin{eqnarray*}
&&f(z)\\
&=&\left \{
\begin{array}{lr}
0, \text{ ~~~~~~~~~~~~~~~~~~~~~if } z < 0,\\
\frac{2}{P(y_x)+P({y'}_{x'})}, \text{ if }  0 \le z < P({y}_{x}),\\
0, \text{ ~~~~~~~~~~~~~~~~~~~~~if } z \ge P({y}_{x}).
\end{array}
\right \},
\end{eqnarray*}
Then, we have,
\begin{eqnarray*}
&&E(U-U')\\
&=&\int_{0}^{P({y}_{x})}\frac{2z}{P(y_x)+P({y'}_{x'})}dz\\
&=&\frac{P^2({y}_{x})}{P(y_x)+P({y'}_{x'})}.
\end{eqnarray*}

We then combine the results of cases $1$ and $3$.\\
Note that when $P(y_x)\ge P({y'}_{x'})$, we have $P({y'}_x)\le({y}_{x'})$,\\
and when $P(y_x) < P({y'}_{x'})$, we have $P({y'}_x)> P({y}_{x'})$,\\
therefore,\\
when $P(y_x)+P({y'}_{x'})\ge 1$, we have,\\
\begin{eqnarray*}
&&E(U-U')\\
&=&\frac{\min\{P^2({y'}_{x}), P^2({y}_{x'})\}}{P({y'}_x) + P({y}_{x'})}.
\end{eqnarray*}
Similarly, after combine the results of cases $2$ and $4$, we have,\\
when $P(y_x)+P({y'}_{x'}) <  1$, we have,\\
\begin{eqnarray*}
&&E(U-U')\\
&=&\frac{\min\{P^2({y}_{x}), P^2({y'}_{x'})\}}{P({y}_x) + P({y'}_{x'})}.
\end{eqnarray*}
Now, note that when $P(y_x)+P({y'}_{x'})\ge 1$, we have,\\
$P(y_x)+P({y'}_{x'})\ge P({y'}_x)+P({y}_{x'})$.\\
Also, $P(y_x)\ge P({y}_{x'})$ and $P({y'}_{x'})\ge P({y'}_{x})$,\\
thus, $\min\{P^2({y}_{x}), P^2({y'}_{x'})\} \ge \min\{P^2({y'}_{x}), P^2({y}_{x'})\}$.\\
Similarly, when $P(y_x)+P({y'}_{x'})< 1$, we have,\\
$P(y_x)+P({y'}_{x'}) < P({y'}_x)+P({y}_{x'})$,\\
and, $\min\{P^2({y}_{x}), P^2({y'}_{x'})\} < \min\{P^2({y'}_{x}), P^2({y}_{x'})\}$.
Thus, finally, we have,
\begin{eqnarray*}
&&E(U-U')\\
&=&\frac{\min\{P^2(y_x), P^2({y'}_x), P^2({y}_{x'}), P^2({y'}_{x'})\}}{\min\{P(y_x)+P({y'}_{x'}),P({y'}_x)+P({y}_{x'})\}}.
\end{eqnarray*}
\end{proof}
\end{reptheorem}

\subsection{Proof of Corollary \ref{thm5}}
First, when $C$ is a set of variables that does not contain any descendant of $X$. We have the conditional on $c$ version of Theorems \ref{thm1} and \ref{thm2} for c-PNS $P(y_x,{y'}_{x'}|c)$ as following. The proof is exactly the same as above, but with every probability conditioned on $c$. 

\begin{corollary}
Given a causal diagram $G$ and distribution compatible with $G$ with experimental data $P(y_x|c)$ and $P(y_{x'}|c)$, let $C$ be a set of variables that does not contain any descendant of $X$ in $G$. let $D=P(y|c)$ be a random variable. Let $L$ be the lower bound of c-PNS using pure experimental data and $L'$ be the lower bound of c-PNS using a combination of experimental and observational data. If $D$ is uniformly distributed on its feasible interval $[max(0, P(y_x|c)-P({y'}_{x'}|c)), min(1, P(y_x|c)+P(y_{x'}|c))]$ ,and $P(y_x|c)+P(y_{x'}|c)\ne 0$ and $P({y'}_x|c)+P({y'}_{x'}|c)\ne 0$, then we have the expectation of the increased lower bound $E(L'-L)$ as follows:
\begin{eqnarray*}
&&E(L'-L)\\
&=&\frac{\min\{P^2(y_x|c), P^2({y'}_x|c), P^2({y}_{x'}|c), P^2({y'}_{x'}|c)\}}{\min\{P(y_x|c)+P(y_{x'}|c),P({y'}_x|c)+P({y'}_{x'}|c)\}}
\end{eqnarray*}
where,\\
\begin{eqnarray*}
P({y'}_x|c) = 1 - P(y_x|c),\\
P({y'}_{x'}|c) = 1 - P(y_{x'}|c),
\end{eqnarray*}
\begin{eqnarray*}
\text{L'} = \max \left \{
\begin{array}{cc}
0, \\
P(y_x|c) - P(y_{x'}|c), \\
D - P(y_{x'}|c), \\
P(y_x|c) - D
\end{array}
\right \},
\end{eqnarray*}
\begin{eqnarray*}
\text{L} = \max \left \{
\begin{array}{cc}
0, \\
P(y_x|c) - P(y_{x'}|c)
\end{array}
\right \}.
\end{eqnarray*}
\begin{proof}
The proof follows exactly the same as Theorem \ref{thm1}, but with every probability conditioned on $c$.
\end{proof}
\label{thm6}
\end{corollary}

\begin{corollary}
Given a causal diagram $G$ and distribution compatible with $G$ with experimental data $P(y_x|c)$ and $P(y_{x'}|c)$, let $C$ be a set of variables that does not contain any descendant of $X$ in $G$. let $D=P(x,y|c)+P(x',y'|c)$ be a random variable. Let $U$ be the upper bound of c-PNS using pure experimental data and $U'$ be the upper bound of c-PNS using a combination of experimental and observational data. If $D$ is uniformly distributed on its feasible interval $[max(0, P(y_x|c)-P(y_{x'}|c)), min(1, P(y_x|c)+ P({y'}_{x'}|c))]$, and $P(y_x|c)+P({y'}_{x'}|c)\ne 0$ and $P({y'}_x|c)+P({y}_{x'}|c)\ne 0$, then we have the expectation of the decreased upper bound $E(U-U')$ as follows:
\begin{eqnarray*}
&&E(U-U')\\
&=&\frac{\min\{P^2(y_x|c), P^2({y'}_x|c), P^2({y}_{x'}|c), P^2({y'}_{x'}|c)\}}{\min\{P(y_x|c)+P({y'}_{x'}|c),P({y'}_x|c)+P({y}_{x'}|c)\}}
\end{eqnarray*}
where,
\begin{eqnarray*}
P({y'}_x|c) = 1 - P(y_x|c),\\
P({y'}_{x'}|c) = 1 - P(y_{x'}|c),
\end{eqnarray*}
\begin{eqnarray*}
\text{U'} = \min \left \{
\begin{array}{cc}
 P(y_x|c), \\
 P(y'_{x'}|c), \\
D, \\
P(y_x|c) + P({y'}_{x'}|c) -D
\end{array}
\right \},
\end{eqnarray*}
\begin{eqnarray*}
\text{U} = \min \left \{
\begin{array}{cc}
 P(y_x|c), \\
 P(y'_{x'}|c)
\end{array}
\right \}.
\end{eqnarray*}
\label{thm7}
\begin{proof}
The proof follows exactly the same as Theorem \ref{thm2}, but with every probability conditioned on $c$.
\end{proof}
\end{corollary}

Now, let us proof Corollary \ref{thm5}.

\begin{repcorollary}{thm5}
Given a causal diagram $G$ and distribution compatible with $G$ with experimental data $P(y_x|c)$ and $P(y_{x'}|c)$, let $C$ be a set of variables that does not contain any descendant of $X$ in $G$. let $D=P(y|c)$ be a random variable, and let $D'=P(x,y|c)+P(x',y'|c)$ be another random variable. Let $LB,UB$ be the lower and upper bound of the benefit function using pure experimental data, respectively. Let $LB',UB'$ be the lower and upper bound of the benefit function using a combination of experimental and observational data, respectively. If $D$ is uniformly distributed on its feasible interval $[max(0, P(y_x|c)-P({y'}_{x'}|c)), min(1, P(y_x|c)+P(y_{x'}|c))]$ and $D'$ is uniformly distributed on its feasible interval $[max(0, P(y_x|c)-P(y_{x'}|c)), min(1, P(y_x)+ P({y'}_{x'}|c))]$, and $P(y_x|c)+P({y'}_{x'}|c)\ne 0$ and $P({y'}_x|c)+P({y}_{x'}|c)\ne 0$ and $P(y_x|c)+P(y_{x'}|c)\ne 0$ and $P({y'}_x|c)+P({y'}_{x'}|c)\ne 0$, then we have the expectation of the increased lower bound $E(LB'-LB)$ and the decreased upper bound $E(UB-UB')$ as follows:
\begin{eqnarray*}
&&E(LB'-LB)\\
&=&\sigma\frac{\min\{P^2(y_x|c), P^2({y'}_x|c), P^2({y}_{x'}|c), P^2({y'}_{x'}|c)\}}{\min\{P(y_x|c)+P(y_{x'}|c),P({y'}_x|c)+P({y'}_{x'}|c)\}}\\
&&\text{~~~~~if }\sigma > 0,\\
&&E(LB'-LB)\\
&=&-\sigma\frac{\min\{P^2(y_x|c), P^2({y'}_x|c), P^2({y}_{x'}|c), P^2({y'}_{x'}|c)\}}{\min\{P(y_x|c)+P({y'}_{x'}|c),P({y'}_x|c)+P({y}_{x'}|c)\}}\\
&&\text{~~~~~if }\sigma < 0,\\
&&E(UB-UB')\\
&=&\sigma\frac{\min\{P^2(y_x|c), P^2({y'}_x|c), P^2({y}_{x'}|c), P^2({y'}_{x'}|c)\}}{\min\{P(y_x|c)+P({y'}_{x'}|c),P({y'}_x|c)+P({y}_{x'}|c)\}}\\
&&\text{~~~~~if }\sigma > 0,\\
&&E(UB-UB')\\
&=&-\sigma\frac{\min\{P^2(y_x|c), P^2({y'}_x|c), P^2({y}_{x'}|c), P^2({y'}_{x'}|c)\}}{\min\{P(y_x|c)+P(y_{x'}|c),P({y'}_x|c)+P({y'}_{x'}|c)\}}\\
&&\text{~~~~~if }\sigma < 0,
\end{eqnarray*}
where,
\begin{eqnarray*}
&&\text{the benefit function~} f(c)=\beta P(y_x,y'_{x'}|c)+\\
&&+\gamma P(y_x,y_{x'}|c)+\theta P(y'_x,y'_{x'}|c) + \delta P(y_{x'},y'_{x}|c),\\
&&\sigma = \beta - \gamma - \theta + \delta,\\
&&W=(\gamma -\delta)P(y_x|c)+\delta P(y_{x'}|c)+\theta P(y'_{x'}|c),\\
&&P({y'}_x|c) = 1 - P(y_x|c),\\
&&P({y'}_{x'}|c) = 1 - P(y_{x'}|c),\\
&&LB = W+\sigma U\text{~~~~~~~~if }\sigma < 0,\\
&&LB = W+\sigma L\text{~~~~~~~~if }\sigma > 0,\\
&&LB' = W+\sigma U'\text{~~~~~~~~if }\sigma < 0,\\
&&LB' = W+\sigma L'\text{~~~~~~~~if }\sigma > 0,\\
&&UB = W+\sigma L\text{~~~~~~~~if }\sigma < 0,\\
&&UB = W+\sigma U\text{~~~~~~~~if }\sigma > 0,\\
&&UB' = W+\sigma L'\text{~~~~~~~~if }\sigma < 0,\\
&&UB' = W+\sigma U'\text{~~~~~~~~if }\sigma > 0,\\
&&\text{L'} = \max \left \{
\begin{array}{cc}
0, \\
P(y_x|c) - P(y_{x'}|c), \\
D - P(y_{x'}|c), \\
P(y_x|c) - D
\end{array}
\right \},\\
&&\text{L} = \max \left \{
\begin{array}{cc}
0, \\
P(y_x|c) - P(y_{x'}|c)
\end{array}
\right \},\\
&&\text{U'} = \min \left \{
\begin{array}{cc}
 P(y_x|c), \\
 P(y'_{x'}|c), \\
D', \\
P(y_x|c) + P({y'}_{x'}|c) -D'
\end{array}
\right \},\\
&&\text{U} = \min \left \{
\begin{array}{cc}
 P(y_x|c), \\
 P(y'_{x'}|c)
\end{array}
\right \}.
\end{eqnarray*}
\begin{proof}
By Corollary \ref{thm6}, we have,\\
\begin{eqnarray*}
&&E(L'-L)\\
&=&\frac{\min\{P^2(y_x|c), P^2({y'}_x|c), P^2({y}_{x'}|c), P^2({y'}_{x'}|c)\}}{\min\{P(y_x|c)+P(y_{x'}|c),P({y'}_x|c)+P({y'}_{x'}|c)\}},
\end{eqnarray*}
and by Corollary \ref{thm7}, we have,\\
\begin{eqnarray*}
&&E(U-U')\\
&=&\frac{\min\{P^2(y_x|c), P^2({y'}_x|c), P^2({y}_{x'}|c), P^2({y'}_{x'}|c)\}}{\min\{P(y_x|c)+P({y'}_{x'}|c),P({y'}_x|c)+P({y}_{x'}|c)\}}.
\end{eqnarray*}
Case 1: $\sigma > 0$.\\
\begin{eqnarray*}
&&E(LB'-LB)\\
&=&E(W+\sigma L' - W -\sigma L)\\
&=&\sigma E(L'-L)\\
&=&\sigma \frac{\min\{P^2(y_x|c), P^2({y'}_x|c), P^2({y}_{x'}|c), P^2({y'}_{x'}|c)\}}{\min\{P(y_x|c)+P(y_{x'}|c),P({y'}_x|c)+P({y'}_{x'}|c)\}},
\end{eqnarray*}
\begin{eqnarray*}
&&E(UB-UB')\\
&=&E(W+\sigma U - W -\sigma U')\\
&=&\sigma E(U-U')\\
&=&\sigma \frac{\min\{P^2(y_x|c), P^2({y'}_x|c), P^2({y}_{x'}|c), P^2({y'}_{x'}|c)\}}{\min\{P(y_x|c)+P({y'}_{x'}|c),P({y'}_x|c)+P({y}_{x'}|c)\}}.
\end{eqnarray*}
Case 2: $\sigma < 0$.\\ 
\begin{eqnarray*}
&&E(LB'-LB)\\
&=&E(W+\sigma U' - W -\sigma U)\\
&=&-\sigma E(U-U')\\
&=&-\sigma \frac{\min\{P^2(y_x|c), P^2({y'}_x|c), P^2({y}_{x'}|c), P^2({y'}_{x'}|c)\}}{\min\{P(y_x|c)+P({y'}_{x'}|c),P({y'}_x|c)+P({y}_{x'}|c)\}},
\end{eqnarray*}
\begin{eqnarray*}
&&E(UB-UB')\\
&=&E(W+\sigma L - W -\sigma L')\\
&=&-\sigma E(L'-L)\\
&=&-\sigma \frac{\min\{P^2(y_x|c), P^2({y'}_x|c), P^2({y}_{x'}|c), P^2({y'}_{x'}|c)\}}{\min\{P(y_x|c)+P(y_{x'}|c),P({y'}_x|c)+P({y'}_{x'}|c)\}}.
\end{eqnarray*}
\end{proof}
\end{repcorollary}
\end{document}